# What makes for an enjoyable protagonist?

# An analysis of character warmth and competence

Hannes Rosenbusch (h.rosenbusch@uva.nl; 0000-0002-4983-3615)
Department of Psychological Methods, University of Amsterdam

Drawing on psychological and literary theory, we investigated whether the warmth and competence of movie protagonists predict IMDb ratings, and whether these effects vary across genres. Using 2,858 films and series from the Movie Scripts Corpus, we identified protagonists via AI-assisted annotation and quantified their warmth and competence with the LLM_annotate package ([1]; human-LLM agreement: r = .83). Preregistered Bayesian regression analyses revealed theory-consistent but small associations between both warmth and competence and audience ratings, while genre-specific interactions did not meaningfully improve predictions. Male protagonists were slightly less warm than female protagonists, and movies with male leads received higher ratings on average (an association that was multiple times stronger than the relationships between movie ratings and warmth/competence). These findings suggest that, although audiences tend to favor warm, competent characters, the effects on movie evaluations are modest, indicating that character personality is only one of many factors shaping movie ratings. AI-assisted annotation with LLM_annotate and gpt-4.1-mini proved effective for large-scale analyses but occasionally fell short of manually generated annotations.

# What makes for an enjoyable protagonist?
# An analysis of character warmth and competence

Fictional characters have distinct personalities [2, 3, 4, 5]. Whether they are friendly and naïve like Forrest Gump or rebellious and brave like Mulan, we relate to them as if they were real people [6, 7, 8, 9]. In fact, literary and cinematic research suggests that our enjoyment of stories hinges primarily on our identification with—or at least our interest in—the principal characters [7, 9]. Here, we investigate the associations between the personality of movie protagonists and movie ratings on IMDb, and how these associations vary across genres.

**Character personality**

Certain traits of fictional characters make it more (or less) likely that we empathize with them [6, 10, 11]. Some studies (and movie reviews) suggest that characters should be "realistic," "complex," or "3D" [12, 13, 14, 15]. However, characters should also remain understandable, meaning their traits and motivations should become readily apparent to the observer [16]. Unclear or incoherent character behavior is usually met with disapproval by audience members [11]. Thus, writers often create characters with recognizable yet extreme personalities that adhere to well-known archetypes [17, 18].

Realism and understandability are important qualities, but they are usually insufficient to engage people in a character's story; there are plenty of realistic and understandable characters that do not hold our interest. Thus, researchers have analyzed which additional features boost the appeal of fiction characters, with a prominent example being dispositional *warmth* [12, 19]. Book reviewers are more likely to endorse a book if they liked the protagonist [20], and people generally enjoy spending time with moral characters [12]. Psychological theory also supports the suggested effect; warmth and agreeableness are crucial traits for social bonding [21]. People are more likely to approach

someone—physically and emotionally—if they appear agreeable, warm, and pro-social [21, 22, 23], which could lay the foundation for enjoying a story about this person [12], especially if the story has a happy ending [24].

Thus, interdisciplinary scholarship supports the Warmth Hypothesis:

*Movies with warmer protagonists receive more favorable audience evaluations.*

However, it is undeniable that many popular stories have fairly disagreeable protagonists [4, 25]. There are countless successful movies about grumpy detectives, psychopathic geniuses, and dismissive love interests. If the next James Bond stopped drinking alcohol, kept to the speed limit, and disarmed villains with kindness and diplomacy—fans would be confused, and movie ratings would likely plummet.

To shed light on people's paradoxical liking of unlikable characters, psychologists have considered complementary traits. They find that low-warmth individuals *can* be liked, as long as they compensate for their lack of warmth with other attributes, most commonly competence [26, 27]. Accordingly, TV show audiences have been found to embrace characters for their warmth or competence, depending on the portrayals [28].

Rather than *liking* competent characters, audiences might admire them for their success, intelligence, or other desirable attributes, and engage in wishful identification [29]. Inhabiting a super competent protagonist allows observers to feel powerful, making for a pleasurable viewing experience [30, 31].

In fact, as soon as characters are admired, their immoral behaviors and social shortcomings are actively justified and excused by fans [32]. Thereby, competent characters can *appear* warmer, more attractive, and more moral to onlookers, especially in collaborative scenarios [33, 34], or when their questionable behaviors are committed for a good cause [35].

Thus, next to agreeableness, we investigate the Competence Hypothesis:

*Movies with more competent protagonists receive more favorable audience evaluations.*

**Genre-specific preferences**

Given the potential benefits of warmth and competence, it may seem optimal to create protagonists with *both* attributes. However, such multi-talented characters are often judged as too perfect, unrealistic, forgettable, and less relatable [25, 36, 37]. Based on reviews and writing tutorials, many people subscribe to the credo that good characters have flaws [38].

Thus, writers often strive for a balance of warmth and competence when creating their characters, either intuitively or deliberately, and the correct choice may depend on the story's target audience [39]. Some viewers enjoy the vicarious rush of danger and skill when watching a criminal mastermind outmaneuver the police in a high-stakes car chase. Others prefer the cozy comfort of a likable protagonist awkwardly bumping into their love interest in a lighthearted romance or comedy film [40, 41, 42]. Considering these different audiences, it therefore stands to reason that a good protagonist aids in fulfilling their respective viewers' needs and desires. A moral and agreeable protagonist can facilitate feelings of warmth and imagined companionship among viewers of a romantic movie [43]. Similarly, a strong and daring hero can help action fans feel empowered and thrilled [44]. In sum, the varying motivations of audience members entail a varying importance of character agreeableness and competence, respectively.

Thus, we also investigate the Genre Hypotheses:

*a) The association between warmth and movie ratings will vary across genres.*

*b) The association between competence and movie ratings will vary across genres.*

# Method

We test the Warmth, Competence, and Genre Hypotheses (see above) by regressing movie ratings on protagonists' warmth and competence inferred from character behaviors in the original movie scripts. The inferences will be collected via the python package LLM_annotate [1], which facilitates character annotations with AI agents and a human-in-the-loop GUI to ensure annotation accuracy. Preregistration, data, and scripts for the current study can be found here:

https://github.com/hannesrosenbusch/character-warmth-competence.

## Data

I analyze the 2,858 movies and series (from here: movies) in the Movie Scripts Corpus dataset from Kaggle [45]. This dataset includes titles, release years, directors, genres, IMDb ratings, and full-length scripts, which will be used to annotate the warmth and competence of the protagonists. Given the potentially high number of genres, we will only analyze the twelve most common genres in the dataset.

## Procedure

I preregistered to use the character with the largest number of dialogue lines (according to the Kaggle character data) as the protagonist. However, this procedure often failed to identify the actual protagonist (e.g., none of the James Bond movies listed James Bond as the protagonist, errors which seemed to originate from incomplete data in the kaggle dataset). Instead, I used a search-enabled LLM (gpt-4.1-mini) to identify the actual protagonist of the movie (see supplementary scripts). Manual inspection of the first 30 protagonists showed 100% agreement with my own protagonist selection. Whenever two major characters had equal importance (e.g., Beavis and Butt-Head), they were both chosen as the movie's protagonist.

Warmth and competence scores were generated with the LLM_annotate package. While listing actions, statements, and prominent omissions of the chosen character in the provided script, the LLM rates each observation on a given trait and the package's default 6-point scale (e.g., "Slaps Mary in the face —> minus two warmth rating"). The characters' overall score consists of the average rating of their actions. LLM_annotate requires a definition and example indicators of the annotated traits, which were set in line with the trait explanations in [51]:

"A character's warmth refers to their tendency to be good-natured, trustworthy, tolerant, friendly, and sincere. Positive examples: Organizes a get-together, admits to insecurity. Negative examples: Rejects someone, boasts."

"A character's competence refers to their tendency to behave capably, skillfully, intelligently, and confidently. Positive examples: Wins something, takes a calculated risk. Negative examples: Embarrasses themselves, fails to understand something."

Human supervision is given by selecting a random sample of 100 scores across movies for manual review in LLM_annotate's GUI. To balance costs and quality, I chose gpt-4.1-mini as LLM_annotate's backend. See supplementary Python scripts for all scripts as well as LLM annotations, and human evaluations.

**Statistical Analysis**

All hypothesis tests will be conducted in R using the brms package [46]. Bayesian regression models will be fitted with the package's default settings, provided that convergence diagnostics are satisfactory. Bayes Factors will be computed against a point null hypothesis. Continuous variables were standardized. I preregistered that I would use stretched beta priors for regression coefficients; however, they are not provided in the brms package, and I therefore chose wide normal priors ($M = 0$, $SD = 1$).

To test the Warmth and Competence Hypotheses, IMDb ratings were regressed on the protagonists' scores in separate models. To test the Genre Hypotheses, I include interaction terms between genres and the respective trait. If these extended models provided a better fit than the main-effect-only model (according to Bayes Factor comparisons), it would serve as evidence in favor of the Genre Hypotheses. As preregistered, I also explored the interaction between warmth and competence, and their respective interactions with gender.

## Results

The distribution of genres, warmth ($M$ = 1.033, $SD$ = 0.575, avg. annotations per character = 183.245), competence ($M$ = 1.07, $SD$ = 0.501, avg. annotations per character = 228.407), and IMDb ratings ($M$ = 6.843, $SD$ = 0.947) is depicted in Figure 1. Warmth and competence were positively correlated ($r$ = .289). Scraping IMDb ratings failed for 13 titles as the website did not display an average rating for these cases (mostly series). Good convergence statistics were achieved for all models presented below (no $\hat{R}$ > 1.01, no deviant transitions after warmup, satisfactory visual inspection of trace plots).

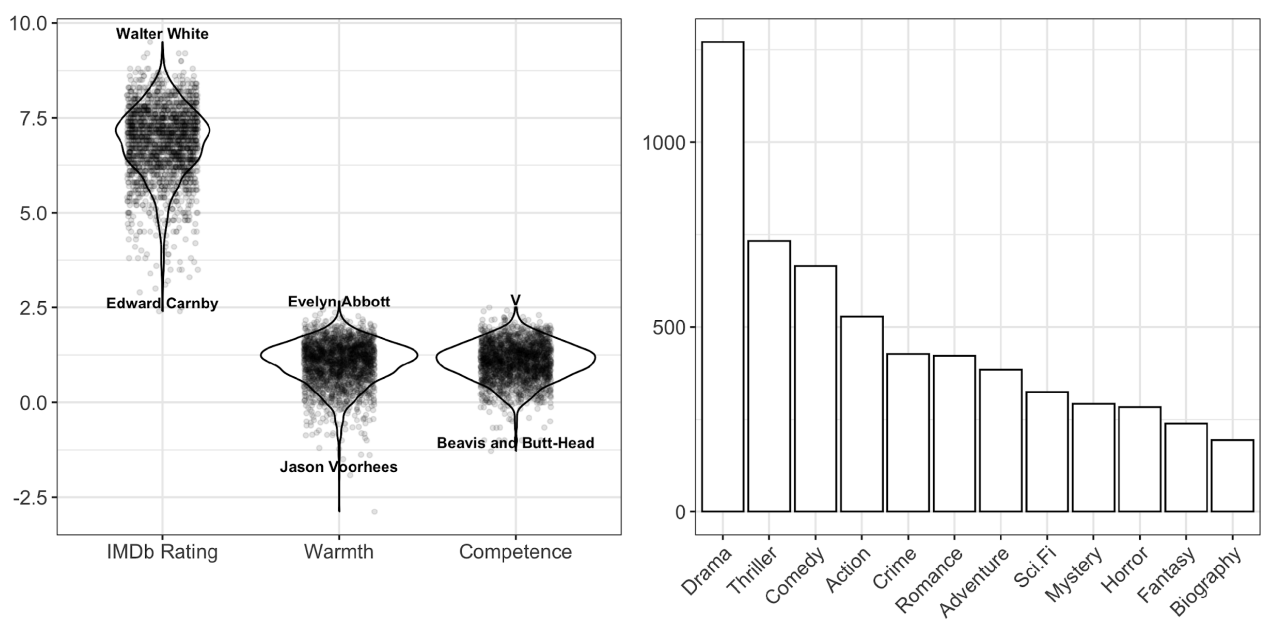

**Figure 1.** Distribution of focal variables before standardization.

**Accuracy of LLM_annotate.** Reviewing 100 randomly sampled LLM annotations revealed a correlation between LLM ratings and human ratings of $r$ = .83. Given that most protagonists have more than a hundred rated annotations, the average LLM rating should therefore be very close to a human-generated character score.

**Warmth hypothesis.** As predicted, regressing IMDb ratings on protagonists' warmth revealed a positive association and a superior model fit than an intercept-only model (BF = 2.2 × 10^7). However, the effect of protagonist warmth was very small ($β$ = 0.04, 95% CI [0, 0.09]). Translated to its original scale, this effect signifies that a movie with a protagonist who is one SD above the average warmth receives an audience score that is 0.042 points higher on IMDb's ten-point scale on average.

**Competence hypothesis.** As predicted, regressing IMDb ratings on protagonists' competence revealed a positive association and a superior model fit than the intercept-only model (BF =1.36 × 10^7). However, the observed effect size was again very small ($β$ = 0.06, 95% CI [0.01, 0.10]). Translated to its original scale, this effect signifies that a movie with a protagonist who is one SD above the average competence receives an audience score that is 0.055 points higher on IMDb's ten-point scale on average.

**Genre hypotheses.** Warmth and competence of protagonists varied across genres, leading to clear improvements over intercept-only models ($BF_{warmth}$ = 6.89 × 10^24; $BF_{competence}$ = 7.52 × 10^83; see Figure 3).

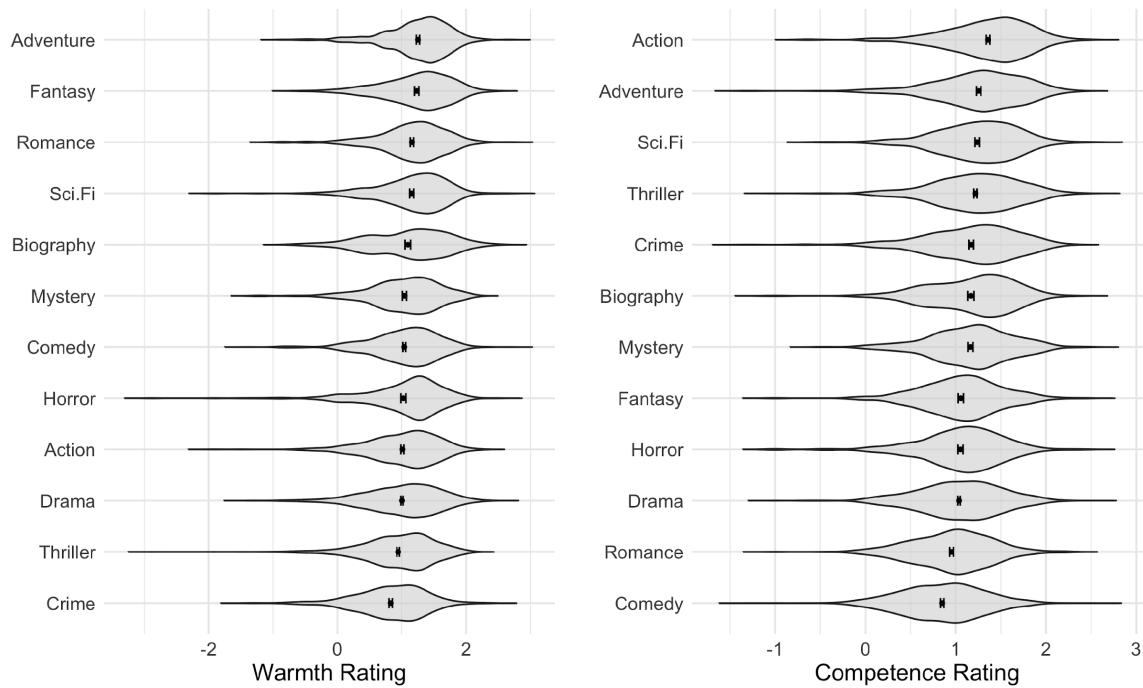

**Figure 3.** Distribution of warmth and competence across genres

When testing the variation of warmth and competence effects on IMDb ratings, the initial multiple-genre-membership models did not converge. To simplify the model structure, each movie was restricted to at most 3 (randomly chosen) genres (which affected 21% of movies). The altered data led to good convergence; however, contrary to the Genre Hypotheses, adding interaction effects between genres and warmth, as well as between genres and competence, led to *worse* model fits ($BF_{warmth}$ = 0.058; $BF_{competence}$ = 0.125). Thus, the effect of protagonist warmth and competence *within* genres did not differ substantially from the trait's average effect. Similarly, the interaction between both traits did not improve model fit beyond the respective main effects (BF = 0.06).

**Protagonist gender.** Male protagonists (N = 1584) were rated as being less warm ($β$ = -0.12, 95% CI [-0.18, -0.06]) and similarly competent ($β$ = 0.04, 95% CI [-0.01, 0.09]) as female protagonists (N = 465). Having a male (as opposed to female) protagonist had a positive association with IMDb ratings ($β$ = 0.36, 95% CI [0.26, 0.47]). Translated to its original scale, this effect signifies that movies with a male protagonist received an audience score that was 0.34 points higher than movies with a female protagonist. The small positive effects of warmth and competence on IMDb ratings were stable across male and female protagonists (comparisons with main-effect-only models: $BF_{warmth}$ = 0.39; $BF_{competence}$ = 0.27).

## Discussion

We are attracted to warm and competent people [51, 47]. Should screenwriters therefore provide us with warm and competent heroes? My analysis suggests that, in line with previous insights about audience identification, there are indeed positive associations between these traits and audience ratings [12, 20]. However, the observed effects are so small that they do *not* warrant a change in the way that writers of popular movies currently design their characters. Even when characters score very high (> 90th percentile) on the respective traits, the relative benefit for movie ratings is more than three times smaller than, for instance, the previously observed effect of protagonist gender (which includes sizeable confounding factors like systematically smaller budgets for women-led movies [48]).

The reasons why protagonists' personalities are only weakly related to movie ratings are relatively intuitive. First, movies can appeal or disappoint in many ways beyond the personality of their protagonists (e.g., other characters, plot, societal relevance, set design, technical implementation, marketing efforts). Second, protagonists themselves can appeal through a wide variety of behaviors, situations, and traits beyond their overall warmth and competence. Third, protagonists can also appeal to audiences because they *change* over the course of a movie ([49], for a review of character arcs, see [50]). Such trait changes are averaged out in the current analysis and likely warrant further exploration.

Overall, this paper does *not* encourage writers to heighten their characters' warmth and competence in formulaic ways, as this would increase the risks of character artificiality, homogeneity, and potential inconsistency, likely offsetting the small benefits of personality finetuning.

At the same time, I do not want to suggest that character attributes do not matter. Audiences often have genre-specific expectations for protagonists (e.g., heroes and villains in adventure stories are usually risk takers, at least sometimes). Accordingly, there was clear evidence that characters from different genres vary in their traits (on average). However, given the absent interaction between genre and personality, the writers in the analyzed sample already adhered to such genre (and gender) specific expectations so that additional warmth and competence revisions are (on average) of little benefit. It would be worthwhile to rerun the current analyses for amateur productions which might be more likely to violate genre-specific character expectations resulting in larger and more genre-dependent effects.

On a methodological subject, the AI-generated annotations were accurate and could be used for a large-scale analysis, but they did not reach the level of human reviewers who can, for instance, reinterpret earlier character behaviors after receiving revealing information at the end of the movie.

## Conclusion

There were small, theory-consistent associations between movie ratings and protagonists' warmth, as well as between movie ratings and protagonists' competence. While character personalities varied across genres, the benefits of warmth and competence stayed relatively stable. Overall, when predicting whether a character is appealing or "well-written", additional variables need to be considered, most notably audience expectations and preferences.